# Bidirectional Heuristic Search Reconsidered

**Hermann Kaindl**                                        HERMANN.KAINDL@SIEMENS.AT
**Gerhard Kainz**                                          GERHARD.KAINZ@SIEMENS.AT
*Siemens AG Österreich, PSE*
*Geusaugasse 17*
*A-1030 Vienna, Austria*

## Abstract

The assessment of bidirectional heuristic search has been incorrect since it was first published more than a quarter of a century ago. For quite a long time, this search strategy did not achieve the expected results, and there was a major misunderstanding about the reasons behind it. Although there is still wide-spread belief that bidirectional heuristic search is afflicted by the problem of search frontiers passing each other, we demonstrate that this conjecture is wrong. Based on this finding, we present both a new generic approach to bidirectional heuristic search and a new approach to dynamically improving heuristic values that is feasible in bidirectional search only. These approaches are put into perspective with both the traditional and more recently proposed approaches in order to facilitate a better overall understanding. Empirical results of experiments with our new approaches show that bidirectional heuristic search can be performed very efficiently and also with limited memory. These results suggest that bidirectional heuristic search appears to be better for solving certain difficult problems than corresponding unidirectional search. This provides some evidence for the usefulness of a search strategy that was long neglected. In summary, we show that bidirectional heuristic search is viable and consequently propose that it be reconsidered.

## 1. Background and Introduction

When a problem is represented as a *state space* graph, *solutions* to such a problem are paths from a given *start* node $s$ to some *goal/target* node $t$. Finding such a solution can be attempted by searching this graph. If the search is guided by heuristic information, it is called a *heuristic search*. Most of the work on heuristic search for problem solving deals with *unidirectional* approaches, that start from $s$ heading towards some node $t$ (see, e.g., Pearl, 1984).

When there is one goal node $t$ explicitly given and the search operators are reversible, *bidirectional search* is possible, which proceeds both in the forward direction from $s$ to $t$ and in the backward direction from $t$ to $s$ (see, e.g., Nilsson, 1980). Strictly speaking, it is not even required that operators have inverses. It is just necessary that for a given node $n$ the set of parent nodes $p_i$ can be determined for which there exist operators that lead from $p_i$ to $n$. Searching backwards means generating parent nodes successively from the goal node $t$ (see, e.g., Russell & Norvig, 1995). In other words, backward search implements *reasoning* about the operators in the backward direction.

As an illustrating example of a class of problems where bidirectional search can be usefully applied, consider finding find a shortest path between two given places $s$ and $t$ using a given map of some city. In case of one-way streets, bidirectional search implements reasoning like the following: "in order to arrive at $t$, some one-way street leading towards $t$ may be used". In a slightly adapted problem class, the cost of driving some street may be different, depending on the driving direction. A steep street to the top of a mountain may serve as an example. Bidirectional search works also correctly in such a case: the backward search implements reasoning in the backward direction but takes account of the cost of driving in the forward direction. More formally, $k_1(m, n) = k_2(n, m)$ is the cost of an optimal path from $m$ to $n$. $k_2$ is used for notational convenience only.[1] All the bidirectional search algorithms dealt with in this paper work correctly under these conditions and do *not* require that the operators are reversible or that the cost of a path is the same in either direction.

Bidirectional search was shown to be more efficient than its unidirectional counterpart when heuristic knowledge is unavailable, but the inverse result was originally found in experiments for bidirectional *heuristic* search by Pohl (1971). Since this kind of search did not work as expected, there was consensus about the conjecture that bidirectional heuristic search is afflicted by the problem of search frontiers passing each other without intersecting. This situation was metaphorically compared by Pohl to missiles that pass each other, and illustrated in a figure that was reprinted by Nilsson (1980, Fig. 2.11). Nilsson conjectured that in such a case the bidirectional search may expand twice as many nodes as would a unidirectional one.

While the original algorithm BHPA proposed by Pohl (1971) may actually show such inefficient performance, the *missile metaphor* is wrong and misleading. We demonstrate that bidirectional heuristic search is actually *not* afflicted by the problem of search frontiers passing each other. The performance of BHPA is much worse than originally expected because of two very different reasons:

1. BHPA's search frontiers typically go *through each other*.

2. The major effort is spent *after* the search frontiers have already met: for finding better solutions than the one found at the first meeting of the search frontiers up to an optimal one; and finally for proving that there is indeed no better solution possible.

The first reason is specific for BHPA and was incidentally resolved by some of the technical improvements introduced in the related algorithm BS* by Kwa (1989). The second issue, however, is also the major obstacle for efficiency of BS* and actually for any bidirectional search algorithm that performs heuristic *front-to-end* evaluations, i.e., evaluations that estimate the minimal cost of some path from the evaluated node on the search front to $t$. Note, that this is the kind of evaluations also performed by typical unidirectional search.

Because of the common belief in the missile metaphor, however, so-called wave-shaping algorithms were developed by de Champeaux (1983), de Champeaux and Sint (1977), and Politowski and Pohl (1984), with the idea to steer the search "wave-fronts" together. In contrast to BHPA and BS*, these algorithms perform *front-to-front* evaluations, i.e., evaluations that estimate the minimal cost of some path from the evaluated node on one search front to the nodes in the opposing front. In fact, these algorithms achieve large reductions in the number of nodes searched compared to algorithms that perform *front-to-end* evaluations. However, they are either excessively computationally demanding, or they have no restriction on the solution quality. Still, where do these reductions in the number of nodes searched using *front-to-front* evaluations come from? After all, the algorithms performing *front-to-end* evaluations do not suffer from the problem of search frontiers passing each other.

---

1. The notation is explained in the Appendix.

In order to answer this important question, let us shortly focus on a common property of heuristic evaluation functions that estimate the minimal cost of some path by applying heuristic knowledge to the *static* information encoded in the state information of the node evaluated. Such static evaluation functions typically evaluate with some error, i.e., the difference between the minimal cost of a path and its heuristic estimate is in most cases greater than zero. An approach to improve the accuracy of a given static evaluation function is to perform a search and to utilize its results. Since this involves dynamic changes, we call it a *dynamic* evaluation function. Dynamic evaluations through bounded look-ahead search were studied in various contexts by Kaindl and Scheucher (1992).

The static evaluation errors are typically smaller for paths with smaller cost, as also observed by Pearl (1984). *Front-to-front* evaluations are therefore typically more accurate than *front-to-end* evaluations. In addition, the costs of the paths from the nodes on the opposing search frontier to $t$ (or $s$, respectively) are known, and so the overall evaluations by the *front-to-front* algorithms are much more accurate than *front-to-end* evaluations. Since the former utilize the results of the search in the opposing direction, we may view this as an approach to dynamically improving heuristic values from the static evaluation function. Due to this asset, wave-shaping algorithms achieve large reductions in terms of nodes generated since they perform *front-to-front* evaluations. However, they are quite expensive in terms of running time (per node examined), which calls for finding an appropriate balance. In fact, Dillenburg and Nelson (1994) as well as Manzini (1995) developed a more recent non-traditional approach to bidirectional search called *perimeter search* that achieves exactly this.

We devised a new and computationally much cheaper approach to dynamic improvements that we call *difference* approach. It utilizes differences of known costs and their heuristic estimates from a given evaluation function to improve other heuristic estimates from this function. This difference approach can be applied in bidirectional heuristic search algorithms that perform heuristic *front-to-end* evaluations. It is exemplified in two new methods for dynamic improvements of heuristic evaluations during search.

We also devised a new approach to bidirectional heuristic search that performs heuristic *front-to-end* evaluations, where these dynamic improvements of heuristic evaluations during search can be embedded efficiently and effectively. This approach is generic in the sense that it encompasses a whole class of (non-traditional) bidirectional search algorithms. As we show in this paper, it can be instantiated for the case of the availability of sufficient memory as well as for the case of limited memory.

Our results from experiments suggest that bidirectional heuristic search can improve on unidirectional heuristic search with respect to both generated nodes and running time (for certain problems of finding optimal solutions). Since the missile metaphor is wrong, bidirectional heuristic search can do so using our approach *without* the very time-consuming *front-to-front* evaluations. So, bidirectional heuristic search is viable and we propose that it be reconsidered.

This paper is organized in the following manner. First, we discuss previous work and present some new theoretical and empirical results on existing approaches to bidirectional heuristic search. Then we describe our new generic approach to non-traditional bidirectional search and two of its instantiations. Thereafter we propose a new approach to dynamically improving heuristic values that is based on differences between known costs and heuristic estimates. After the presentation of experimental results from applying these approaches, we discuss them in the context of the various approaches to bidirectional heuristic search that were previously proposed.

## 2. Previous Work

In order to make this paper self-contained, we sketch here the essentials of previous work on heuristic search algorithms with a focus on bidirectional heuristic search, without going into more detail than necessary to understand both our new results on this previous work and our new approaches.

### 2.1 Unidirectional Heuristic Search Algorithms

Many unidirectional search algorithms have been presented, so it would be prohibitive to review all of them here. Rather, we focus on those unidirectional algorithms that form the basis of bidirectional search as discussed in this paper. First, we review the traditional *best-first* search algorithm A* (Hart, Nilsson, & Raphael, 1968). Then, we shortly explain the linear-space algorithm IDA* (iterative-deepening-A*) proposed by Korf (1985). Finally, we review an algorithm called *Trans* (Reinefeld & Marsland, 1994) that implements a form of enhanced iterative-deepening search.

A* maintains the set OPEN of so-called *open* nodes that have been generated but not yet expanded, i.e., the frontier nodes. Much as any best-first search algorithm, it always selects a node from OPEN with minimum estimated cost, one of those it considers "best". This node is expanded and moved from OPEN to CLOSED. A* specifically estimates the cost of some node $n$ with an evaluation function of the form $f(n) = g(n) + h(n)$, where $g(n)$ is the (sum) cost of a path found from $s$ to $n$, and $h(n)$ is a heuristic estimate of the cost of reaching a goal from $n$, i.e., the cost of an optimal path from $s$ to some goal $t$. If $h(n)$ never overestimates this cost (it is said to be *admissible*) and if a solution exists, then A* is guaranteed to return an optimal (minimum-cost) solution (it is also said to be *admissible*).

Under certain conditions, A* is optimal over admissible unidirectional heuristic search algorithms using the same information, in the sense that it never expands more nodes than any of these (Dechter & Pearl, 1985). We emphasize here that this optimality result of A* only compares it with *unidirectional* competitors, so a *bidirectional* approach may well improve on the performance of A*. The major limitation of A* is its memory requirement, which is proportional to the number of nodes stored and therefore in most practical cases exponential.

IDA* was designed to address this memory problem, while using the same heuristic evaluation function $f(n)$ as A*. IDA* performs iterations of *depth-first* searches. Consequently, it has linear-space requirements only. Although performing depth-first searches iteratively deeper and deeper has been heavily used in computer chess programs in the context of *alpha-beta minimax* search since the sixties and is still in use (see Kaindl, 1990), the application of this approach in problem-solving searches marked a breakthrough for solving more difficult problems. IDA*'s depth-first searches are guided by a threshold that is initially set to the estimated cost of $s$; the threshold for each succeeding iteration is the minimum $f$-value that exceeded the threshold on the previous iteration.

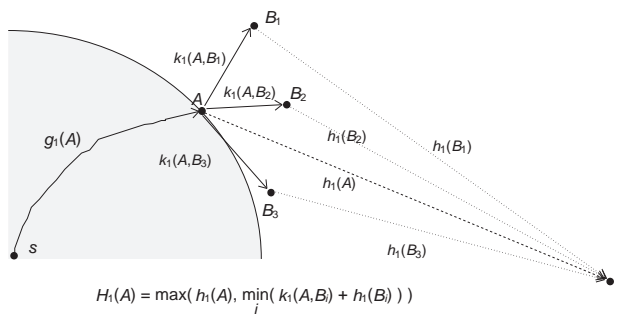

Figure 1: An illustration of the *back-up* idea.

While IDA* shows its best performance on trees, one of its major problems is that in its pure form it cannot deal with duplicate nodes in the sense of *transpositions*. A transposition arises, when several paths lead to the same node, and such a search space is represented by a *directed acyclic graph* (DAG). This disadvantage of IDA* relates to its advantage of requiring only linear space.

Fortunately, most computers have more memory available than needed for IDA*. This memory can be utilized for recognizing duplicate nodes in two ways, using a *finite state machine* (Taylor & Korf, 1993), or a *transposition table* implemented as a hash table (Reinefeld & Marsland, 1994). Due to its more general applicability in a wider variety of domains, and since our bidirectional algorithms partly make use of it, we focus on the latter technique.

The algorithm Trans proposed by Reinefeld and Marsland (1994) uses a transposition table for IDA*. Since the size of such a table can be deliberately parameterized, this is an approach to utilizing limited memory. Analogously to earlier applications of transposition tables in computer chess programs, Trans utilizes its table actually for two purposes:

- for recognizing transpositions;

- for caching the best heuristic values acquired dynamically.

Since the latter use is more difficult to understand, we explain its underlying idea in more depth. This *back-up* idea is illustrated in Fig. 1. When during the normal search the nodes $B_i$ are statically evaluated but not stored, these values can still be used by backing them up to some node $A$ that is stored — in the case of Trans in its transposition table. The *dynamic* value of $A$ is the minimum of the estimated costs of the best paths found through the nodes $B_i$. Unless the static evaluator is consistent, it is useful to store the maximum of the dynamic and the static value of a node. When such a cached node is re-searched, an improved value can often be used instead of the value assigned directly by the static evaluation function.

Apart from its use in Trans, this back-up idea is actually widely applied in many algorithms like MA* (Chakrabarti, Ghose, Acharya, & DeSarkar, 1989), MREC (Sen & Bagchi, 1989), RTA* (Korf, 1990), SMA* (Russell, 1992) and ITS (Ghosh, Mahanti, & Nau, 1994).

Its advantages are very little overhead and steady (though often modest) improvement with increasing memory size. In addition, this idea also works when a goal condition instead of a goal node is specified, i.e., it does not require that a goal node is explicitly given. However, it is only applicable for re-searched and cached nodes, and we cannot see how it could make sense in the context of traditional best-first search like A*.

## 2.2 The Traditional Approach to Bidirectional Heuristic Search

First, we look at the older approach to bidirectional heuristic search where forward and backward searches alternate. We call this the traditional approach. It encompasses both algorithms performing *front-to-end* and others performing *front-to-front* evaluations.

### 2.2.1 FRONT-TO-END EVALUATIONS

Since the first proposed algorithm on bidirectional heuristic search called BHPA (Pohl, 1971) performed *front-to-end* evaluations, let us begin with this approach. It employs heuristic evaluation functions $h_d(n)$ that estimate the cost of an optimal path from the evaluated node $n$ to $t$ or $s$, respectively, depending on the search direction $d$. More precisely, $h_1(n)$ estimates the cost of an optimal path from $n$ to $t$ in the forward search, and $h_2(n)$ from $s$ to $n$ in the backward search. Note, that always an optimal path from $s$ to $t$ is to be found (i.e., *not* from $t$ to $s$) and therefore also the cost of such a path is estimated by the evaluation function $f_d$ that uses $h_d$ as its heuristic component. From the viewpoint of the backward search that targets node $s$, however, it may seem that the cost from its frontier to $s$ is estimated heuristically, while it is more precisely the cost from $s$ to the frontier. This issue matters when the cost of some path is *not* the same in either direction.

We can view a BHPA search essentially as two A*-type searches in opposite directions, i.e., traditional *best-first* searches.[2] These are performed quasi-simultaneously, i.e., on a sequential machine one node is expanded after another, but the search direction is changed (at least) from time to time. The decision for searching in the forward or backward direction is made anew for each node expansion according to the *cardinality criterion* (Pohl, 1971):

**if** $|\text{OPEN}_1| \leq |\text{OPEN}_2|$ **then** $d \leftarrow 1$ **else** $d \leftarrow 2$

Whenever the search frontiers meet at some node $n$, a solution is found. Its cost is $g_1(n) + g_2(n)$, i.e., the cost of the path found by the forward search from $s$ to $n$, plus the cost of the path found by the backward search from $n$ to $t$. Even when the two parts of such a solution of the forward and the backward search are optimal, however, the concatenated solution path is not necessarily optimal. Therefore, such an algorithm requires a special termination condition for guaranteeing optimal solutions. The termination condition of

---

2. More precisely, BHPA can be viewed to consist of two HPA searches (Pohl, 1970) in opposing directions. As long as the heuristic function used is consistent and its values are weighted equally as the $g_d$-values, the only relevant difference is a check whether OPEN has become empty. For *admissible* but not *consistent* heuristic functions, the option to move nodes back from CLOSED to OPEN is important, if a new better $g_d$-value is found. A heuristic function is *consistent* if $h_d(m) \leq h_d(n) + k_d(m, n)$ for all nodes $m$ and $n$. This implies that $h_d$ is *admissible*, i.e., the heuristic function never overestimates the real cost.

BHPA is as follows:

$$L_{min} \leq \max[\min_{x \in \text{OPEN}_1} f_1(x), \min_{x \in \text{OPEN}_2} f_2(x)] \qquad (1)$$

This condition essentially means that the cost $L_{min}$ of the best (least costly) complete path from $s$ to $t$ found so far is not larger than an estimate computed from the $f_d$-values in both search frontiers. If the heuristic used for these estimates is admissible, this path must already be an optimal solution in order to satisfy this termination condition. Since understanding this condition is important for this paper, we elaborate it in more depth below.

Implicitly this is also the condition for successful termination of the improved algorithm BS* (Kwa, 1989), which removes all nodes $n$ whose $f_d$-values are $\geq L_{min}$ and terminates when $\text{OPEN}_1$ or $\text{OPEN}_2$ is empty. This technique of removing nodes is called *trimming* in BS*, and such newly generated nodes are not placed into the sets of open nodes at all, which is called *screening*. While these techniques improve on BHPA "just" with respect to saving memory, BS* additionally includes improvements that reduce the number of nodes generated. These major improvements are the following:

- *nipping*: if a node is selected for expansion which is already in CLOSED in the opposite search tree, it can just be put into CLOSED in the current search tree *without* expansion;

- *pruning*: in the same situation, descendants of this node in OPEN in the opposite search tree can be removed.

Both BHPA and BS* are *admissible* if $f_d$ is *consistent*. However, BHPA's results were clearly less efficient than those of A* for finding optimal solutions, and also BS* was never shown to be really more efficient than A*.

Köll and Kaindl (1993) were the first to conjecture that the missile metaphor is misleading as an explanation and provided some (preliminary) evidence for this finding. Based on it and realizing that fulfilling the termination condition (1) is a key issue, they developed efficient $\varepsilon$-admissible search algorithms, that typically find solutions with a known error bound faster and generate fewer nodes than a corresponding derivative of A* that guarantees the same error bound. These algorithms provided, however, no improvements for finding optimal solutions, and they require exponential space like BHPA, BS* and A*.

Based on the same approach, Kaindl and Khorsand (1994) showed that bidirectional heuristic search using limited memory is possible through using a unidirectional search algorithm that can cope with limited memory — SMA* (Russell, 1992). However, the runtime efficiency was insufficient.

### 2.2.2 Front-to-front Evaluations

Since for a long time there was consensus about the belief that the search frontiers would pass each other, research focused on algorithms that would force the "wavefronts" to meet through "wave-shaping" techniques: BHFFA (de Champeaux & Sint, 1977), BHFFA2 (de Champeaux, 1983), d-node retargeting (Politowski & Pohl, 1984) and a generalized algorithm (encompassing BHPA and BHFFA2) (Davis, Pollack, & Sudkamp, 1984). These algorithms perform *front-to-front* evaluations and they show that bidirectional heuristic search can be efficient in terms of the number of nodes generated.

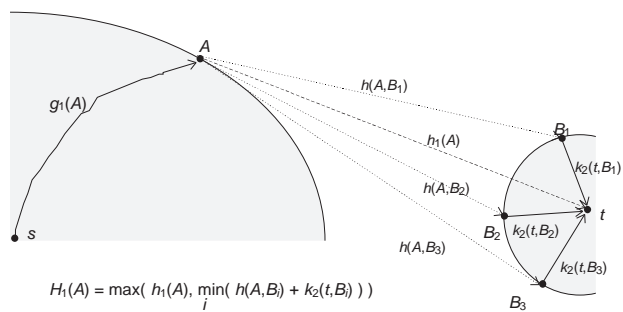

Figure 2: An illustration of the *front-to-front* idea.

Since the basic idea of *front-to-front* evaluations is important for understanding this paper, we illustrate it using Fig. 2. When for the evaluation of some node $A$ the nodes $B_i$ on the opposite search front are available in storage, the costs of optimal paths from $A$ to every $B_i$ can be estimated. Adding these to known costs of paths from $B_i$ to the goal node $t$, normally more accurate dynamic estimates can be gained than from a static *front-to-end* evaluator that directly estimates the cost from $A$ to $t$.

However, the algorithms performing *front-to-front* evaluations are either excessively computationally demanding, or they have no restriction on the solution quality. They have to compute heuristic estimates between all nodes in one search frontier and all nodes in the other, in order to estimate all the paths going through all nodes in the opposite frontier and vice versa. So, their effort for the evaluations needed for a single node selection and expansion may even seem to be proportional to the cross product of the numbers of nodes in both frontiers. Through the use of appropriate data structures, this effort can be reduced to become proportional to the number of descendants of the expanded node times the size of the opposite search frontier.[3] Still, this is excessively computationally demanding for frontiers that may contain in the order of millions of nodes. For keeping this effort practical for non-trivial problems, such an algorithm may either restrict this computation to a certain (small) number of nodes with promising values or keep the search direction focused on a single target node of the opposing frontier for several steps before retargeting it. Both approaches typically terminate with non-optimal solutions and therefore obviously lose admissibility, i.e, the guarantee for finding optimal solutions.

## 2.3 The Non-traditional Approach to Bidirectional Heuristic Search

So, the traditional approaches did not succeed to improve on unidirectional search for finding and guaranteeing optimal solutions. In particular, all these algorithms are based on traditional best-first search that has exponential storage requirements. It may seem that bidirectional search needs to store nodes of at least one frontier so that the search from the opposing side can recognize meeting this frontier (typically implemented through some hashing scheme). Instead of storing both frontiers for having forward and backward searches alternate, it is possible to search in one direction first storing nodes, and then to search in the other direction. We call this here the non-traditional approach to bidirectional heuristic search.

---

3. According to personal communication with Dennis de Champeaux.

Such an approach is the *perimeter search* (Dillenburg & Nelson, 1994; Manzini, 1995). In perimeter search, a breadth-first search generates and stores all the nodes around $t$ up to a predetermined (and fixed) *perimeter depth*. The final frontier of this breadth-first search is called *perimeter*. After this search is finished and the nodes are stored, a forward search starts from $s$, targeting all of the perimeter nodes. Depending on the given problem and the available storage, this forward search can be performed in an A* or IDA* fashion. The former is implemented in PS* (Dillenburg & Nelson, 1994), and the latter both in IDPS* (Dillenburg & Nelson, 1994) and in BIDA* (Manzini, 1995). For the same perimeter depth, IDPS* and BIDA* search exactly the same nodes. However, BIDA* temporarily removes from the perimeter the nodes that cannot affect the computation of its evaluation function and consequently reduces the number of heuristic *front-to-front* evaluations compared to IDPS*. Due to this improvement, BIDA* is far more efficient in terms of running time than IDPS*.

BIDA* achieves very good results in the (sliding-tile) Fifteen Puzzle domain. We investigate below why this is the case in contrast to the traditional approaches to bidirectional heuristic search. In particular, we show the results of experiments with varying perimeter depth, i.e., varying perimeter size or storage use.

## 3. Some New Results on the Previous Approaches

Still, it seems that these previous approaches to bidirectional heuristic search are not understood properly. Therefore, we present some new results about them before we propose our new approaches.

### 3.1 Theoretical Results

We present some new theoretical results on bounds on the number of nodes expanded by traditional bidirectional heuristic search with *front-to-end* evaluations. Since its runtime performance is proportional to the number of nodes expanded, these are bounds on the potential efficiency. We assume the availability of a consistent heuristic evaluation function $h_d$ in both directions.

First we make explicit a principally known result in the form of a lemma, since we need this particular result for proving the new results. In addition, understanding it is important for understanding these results. Note, however, that this termination condition for bidirectional search is significantly different from termination conditions for unidirectional search like A* as given by Pearl (1984).

**Lemma 3.1** (a sufficient condition for successful termination of *BHPA* or *BS\**):

*If there is a solution path from $s$ to $t$, BHPA or BS\* terminate successfully (i.e., by finding such a path) iff both the following conditions are satisfied:*

*(i) in at least one of the search frontiers $d$ of BHPA or BS\* the minimum $f$-value must have been raised at least to the value of an optimal solution $C^*$, that is, $\min_{x \in \text{OPEN}_d} f_d(x) \geq C^*$; and*

*(ii) an optimal solution must have been found, that is, $L_{\min} = C^*$.*

**Proof:** We need not be concerned about whether these algorithms indeed find optimal solutions, since the corresponding proofs were given by Pohl (1971) and Kwa (1989), respectively. We only focus here on how exactly the termination condition in Formula (1) is fulfilled — for BHPA this is the explicit termination condition, for BS* it is implicit as explained above. The minimum $f$-values of $\text{OPEN}_d$ are at first the values $f_1(s)$ and $f_2(t)$, respectively. Since $f_d$ is consistent they do not exceed $C^*$. The minimum $f$-values of $\text{OPEN}_d$ increases only gradually until all the nodes with $f$-values $< C^*$ of at least one search frontier are expanded (or nipped or pruned by BS*). Since the maximum of the minimum $f$-values of $\text{OPEN}_d$ is used, only one but at least one of them must become $\geq C^*$. During the search, $L_{\min} \geq C^*$ always holds, and when an optimal solution is found, $L_{\min} = C^*$. □

In order to establish bounds on the number of node expansions, let us first focus on an upper bound on the number of nodes expanded by BHPA.

**Theorem 3.1** *The number of node expansions of BHPA can be bounded from above by*

$$\#(BHPA) < \#(A^*)_1 + \#(A^*)_2$$

**Proof:** In the worst case, BHPA may have to perform its A*-type searches in both directions completely, with the exception of at least one node expansion. Even when $L_{\min} = C^*$ is achieved only in the last node expansion in one direction, immediately thereafter the termination condition is fulfilled according to Lemma 1. Therefore, in the opposite direction at least one node expansion can be saved. □

In some sense, this bound may look quite weak, but actually Nilsson (1980) conjectured that a bidirectional heuristic search may expand twice as many nodes as would a corresponding unidirectional one. This conjecture was based on the assumption originally published by Pohl (1971) that the search frontiers may pass each other without intersecting.

More recently, however, some empirical evidence was found by Köll and Kaindl (1993) that this assumption is invalid, i.e., the frontiers typically meet rather early even without using wave-shaping techniques. So, the question may arise as to whether and under which conditions the result of Theorem 3.1 is reasonable and useful. In order to show such conditions, we define a strong symmetry property of search spaces. Although this may seem to be a completely unrealistic assumption, it is not too difficult to imagine a search space with this property. Searches for optimal solutions to TSP (traveling salesman problem) instances need to generate nodes that represent visiting all the neighboring cities of the start city. Since this same city is also the final city to be visited, a reverse search in the opposing direction needs to generate nodes for exactly the same cities, etc. So, at least a straight-forward implementation of bidirectional search for the TSP works in a symmetric space. For symmetric TSP instances (where the arc costs are the same independent of the direction) and for usual heuristic evaluations functions for the TSP (like the minimum spanning tree heuristic), it turns out to be a *perfectly A\*-symmetric* search space.

**Definition 3.1** *Let $f_1^1 = h_1(s), f_1^2, \ldots, f_1^{k-1}, f_1^k = C^*$ be the different $f$-values of expanded nodes in the forward direction and analogously $f_2^1 = h_2(t), f_2^2, \ldots, f_2^{k-1}, f_2^k = C^*$ in the backward direction. A search space is* perfectly A\*-symmetric *iff A\* expands the same number of nodes for each $f$-value in the forward direction as in the backward direction, that is, $\#^j(A^*)_1 = \#^j(A^*)_2$ for each $j = 1 \ldots k$.* □

**Theorem 3.2** *If the search space is* perfectly A\*-symmetric *and the f-values are all distinct in each direction, then*

$$\#(BHPA) = 2 \cdot \#(A^*) - \delta \ \ with \ 1 \leq \delta \leq 3$$

**Proof:** In a perfectly A\*-symmetric search space, the numbers of nodes expanded in both directions by the A\*-type searches within BHPA is strictly the same up to the last but 2 $f$-values, because no termination is possible up to this point; and since these are all distinct in each direction, this amounts to 2 nodes each for the remaining 2 $f$-values:

$$\#(A^*)_1 - 2 = \#(A^*)_2 - 2$$

Depending on when $L_{min} = C^*$ is achieved, 1 up to 3 more nodes must be expanded to fulfill the termination condition. Summing up proves the theorem. □

Since in practice the $f$-values are normally not all distinct (in each direction), we show the consequence of a more realistic assumption — the occurrence of many different $f$-values. This is meant in the sense that the number of nodes with the same $f$-value is small compared to the number of nodes expanded.

**Corollary 3.1** *If the search space is* perfectly symmetric *and there are many different f-values, then*

$$\#(BHPA) \approx 2 \cdot \#(A^*)$$

**Proof:** Since there can be several nodes with the same $f$-value, the expansion of more than 3 nodes may be saved when an optimal solution has already been found. Because the number of nodes with the same $f$-value is small compared to the number of nodes expanded, however, $\delta \ll \#(BHPA)$. □

So, under this strong assumption on symmetry BHPA expands close to twice as many nodes as A\*. How is it possible that this conjecture of Nilsson (1980) is supported although its original assumption appears not to be valid?

The point is that the search frontiers of BHPA meet early, i.e., they do *not* pass each other without intersecting, but they go *through each other*! So, there is a possibly large region of the search space explored twice (as illustrated in Fig. 3).

BS\* avoids such double exploration (see again Fig. 3). Unfortunately, it appears to be difficult to quantify the size of this region. So, we cannot determine a tighter upper bound on the number of nodes expanded by BS\* without further assumptions.

Fig. 3 also illustrates that the search frontiers of BS\* are typically "ragged". This means that the meetings occur in the "middle" as well as near $s$ or $t$ (as observed in our experiments).

Now let us have a look at *lower* bounds on the number of nodes expanded by BHPA. We do not need the assumption on symmetry here but we can show more general results.

**Theorem 3.3** *The numbers of nodes expanded by BHPA can be bounded from below by*

$$\min(X_1, X_2) + 1 \leq \#(BHPA)$$

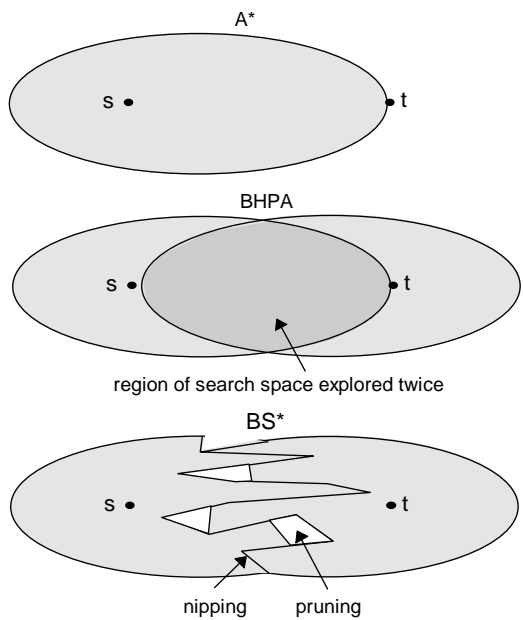

Figure 3: An illustration of traditional bidirectional heuristic search with *front-to-end* evaluations.

where $X_d = \#_d(A^*) - \#_d^k(A^*)$ is the number of nodes that $A^*$ would expand in search direction $d$ minus the number of nodes with value $f_d^k = C^*$.

**Proof:** This lower bound represents the case of earliest termination according to Lemma 1. (At least 1 node is expanded in each direction.) □

**Corollary 3.2** *If the f-values are all distinct in each direction, then the number of nodes expanded by BHPA can be bounded from below by*

$$\min(\#_1(A^*), \#_2(A^*)) \leq \#(BHPA)$$

**Proof:** $X_d = \#_d(A^*) - 1$ since there is only 1 node $n$ with $f_d(n) = C^*$. □

**Corollary 3.3** *The maximal improvement of BHPA over $A^*$ is given by*

$$\#(A^*) - \min(X_1, X_2) - 1.$$

**Proof:** $\min(X_1, X_2) + 1$ is the minimum number of nodes expanded by BHPA. □

In essence, we have shown that under certain conditions traditional bidirectional heuristic search with *front-to-end* evaluations as exemplified by BHPA can expand close to twice as many nodes as $A^*$. While the original conjecture for such a result was based on an apparently wrong assumption, we found that another — even more obvious — effect is (partly) responsible.

In addition, we have shown that BHPA cannot be much more efficient than $A^*$ with respect to node expansions even in the best case. For a variant of BS* without the *pruning* technique, the same lower bound on the number of nodes expanded applies. In general, the major problem of traditional bidirectional heuristic search with *front-to-end* evaluations is the cost of satisfying the termination condition.

### 3.2 Empirical Results

In order to provide evidence that the missile metaphor is misleading, we present some new empirical data on the performance of BS*. Since perimeter search seems to become more and more efficient with increasing perimeter depth (Manzini, 1995), we have investigated its behavior through experiments in two different domains. We present new empirical results from these experiments and provide some explanation why perimeter search works so well in the Fifteen Puzzle domain.

#### 3.2.1 BS*

BS* is a classical best-first search algorithm and requires exponential memory. So, we are not aware of any BS* implementation yet that is able to solve difficult problem instances of the Fifteen Puzzle, given no domain-specific knowledge about the puzzle other than the Manhattan distance heuristic. In our experiments, BS* was able to solve 59 of the 100 instances used by Korf (1985), having available up to 256 Mbytes of main storage (on a Convex C3220).

We gathered some data during these runs of BS* which provide empirical evidence that the missile metaphor is misleading (in addition to the data already given by Köll and Kaindl (1993)). In the average, BS* found the first solution after the generation of 7.2 percent of the total number of nodes generated. The quality of this solution is on average just 6.3 percent worse than that of an optimal solution. After continuing its searches, BS* found optimal solutions after the generation of 22.4 percent of the total number of nodes generated (again on average). That is, most of the search effort of BS* was spent to verify optimality.

That means that the search frontiers of BS* meet relatively early *without* the use of wave-shaping techniques, and even optimal solutions are found rather quickly. However, even when BS* has already found an optimal solution to some problem instance, it does not "know" that this solution is optimal. So, it must continue the search and generate the remaining nodes in order to prove that there is in fact no better solution available.

Relatively to its overall higher effort, BHPA would find a first solution even "earlier" than BS*. Of course, BHPA needs exactly the same number of nodes as BS* for having its search frontiers meet. After this first meeting, however, it would have to generate more nodes than BS* because its search frontiers go through each other. If the search frontiers would, however, pass each other as illustrated by the missile metaphor, solutions could not be found that early.

#### 3.2.2 Perimeter Search

Perimeter search achieved very good results in the Fifteen Puzzle domain, where it can solve any Fifteen Puzzle problem instance relatively fast and with limited memory. However, this approach to bidirectional heuristic search also seems not to be understood sufficiently yet. So, we made experiments with increasing perimeter depth in two different domains. The results may seem to be quite surprising. While we cannot yet explain them theoretically, they are important in their own right, and we try to explain them intuitively.

For these experiments, it was feasible to use the complete set of 100 Fifteen Puzzle problem instances as used by Korf (1985). Fig. 4 shows that in this domain BIDA* works very well, especially in terms of the number of nodes generated. The data are normalized to the respective search effort of IDA* (in Korf's implementation), since it was the first algorithm able to solve random instances of the Fifteen Puzzle.[4] Also the running times are very good.[5]

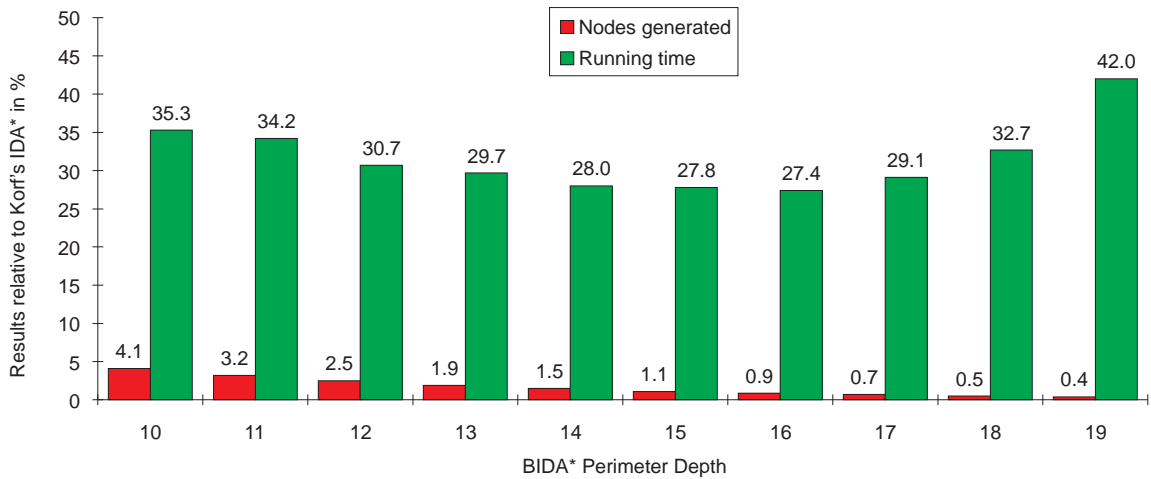

Figure 4: Comparison of BIDA* for different perimeter depths on the Fifteen Puzzle (100 instances) — time optimum.

Consistently with (Manzini, 1995, Table 1), Fig. 4 shows a steady decrease of both the number of nodes generated and the required running time for increasing perimeter depth until it reaches 16. At this perimeter depth, however, BIDA* achieves its minimum running time. The exact perimeter depth where such an optimum occurs may depend on several factors such as the machine used and the efficiency of the implementation. The new and important finding is, however, *that* such an optimum actually exists for BIDA*. While an optimum perimeter depth was shown to exist for PS* by Dillenburg and Nelson (1994), the data presented by Manzini (1995) suggested that for increasing perimeter depth the number of evaluations performed by BIDA* even decreases. For larger perimeter depths, however, the savings in terms of node generation are obviously outweighed by the larger cost of the *front-to-front* evaluations. Note, that the data presented by Manzini (1995) did not show this optimum because of the amount of memory required for storing the perimeter for depths greater than 14 that exhausted the resources available for the experiments reported there.

---

4. Just to give an idea of the overall difficulty of the given problem set, note that IDA* generates some 363 million nodes on average, which needs slightly less than half an hour on a Convex C3220.

5. BIDA*'s result here is worse than the data reported by Manzini (1995). This is primarily due to the use of a different machine and a different implementation that is based on the very efficient code of IDA* for the puzzle provided to us by Korf that we are using. In such an implementation the overhead especially of wave shaping shows up more clearly even when using the runtime optimizations described by Manzini (1995). While we had no access to the implementation by Manzini, in E-mail communication with him we were given some hints about it, and there was agreement about the overall effect on the relative running times due to the different implementations of IDA*.

Knowing about the existence of such an optimum helps us better understand the improvement of perimeter search over the traditional approach to bidirectional heuristic search based on *front-to-front* evaluations as exemplified, e.g., by BHFFA. The advantage of improved evaluation accuracy is to be balanced with the large overhead in time consumption for node evaluations. While BIDA* can be tuned towards this optimum, an algorithm like BHFFA is typically out of balance in this regard. While BHFFA can for this reason only find optimal solutions to quite easy problems, perimeter search is comparably much cheaper per node searched, since a much smaller frontier is "targeted".

Although the performance of perimeter search cannot be improved deliberately through using more and more memory, the optimum running time of BIDA* for the Fifteen Puzzle problems is very good. So, we wanted to see whether and how such results can also be achieved in another domain which we used for experimenting with our own algorithms. We made experiments of finding optimal solutions to a set of maze problems.[6] For these problems, BIDA* based on IDA* is inefficient due to the high number of iterations. So, we used here PS* (Dillenburg & Nelson, 1994) which implements the common underlying idea — perimeter search — based on A*. While A* works very well for such maze problems, it seems that the runtime optimization of BIDA* cannot be practically used in an A*-based algorithm due to excessive storage requirements, since for every node in OPEN information about every perimeter node would have to be stored that may affect the computation of the *front-to-front* evaluations. In fact, Manzini (1995) only states that his technique can be applied to any *depth-first* search algorithm.

Based on these experiments, the perimeter search approach appears not to work satisfactorily as illustrated in Fig. 5 — neither in terms of generated nodes nor in terms of running time. The data are normalized to the respective search effort of A*, since it seems to be the most efficient algorithm for such problem instances that fit into memory (see also the optimality result of A* over unidirectional competitors by Dechter & Pearl, 1985).[7] Even for comparably larger perimeter depths (50, 100, ..., 250), the numbers of generated nodes only marginally improve (up to 93.9 percent of the number of nodes generated by A* as shown in this figure), while the running time becomes quite high (up to 358.7 percent). The running time can be reduced for perimeter depths smaller than 25, but for these no real savings in the number of nodes generated and therefore no improvement over A* can be observed.

When considering these very different performances of perimeter search in these domains, the question arises, *why* it works so well for the Fifteen Puzzle and why not satisfactorily for the maze. Let us consider a reason for the good results first, having a closer look at the case of perimeter depth 1. This minimal perimeter around the node $t$ in the Fifteen

---

6. Our use of this domain was inspired through its use by Rao *et al.* (1991). Problem instances in this domain model the task of navigation in the presence of obstacles. 100 instances were drawn randomly using the approach behind the Xwindows demo package Xmaze. As a heuristic evaluator, we use the Manhattan distance like Rao *et al.* (1991).

For our experiments, we made the following adaptations. In order to allow transpositions, we do not "install" a wall in three percent of the cases. This leads to roughly the same "density" of transpositions as in the Fifteen Puzzle. Moreover, we use much larger mazes — 2000 × 2000, and in order to focus on the more difficult instances of these, we only use instances with $h_1(s) \geq 2000$.

7. Just to give an idea of the overall difficulty of the given problem set, note that A* generates some 2.7 million nodes on average, which needs less than two minutes on a Convex C3220.

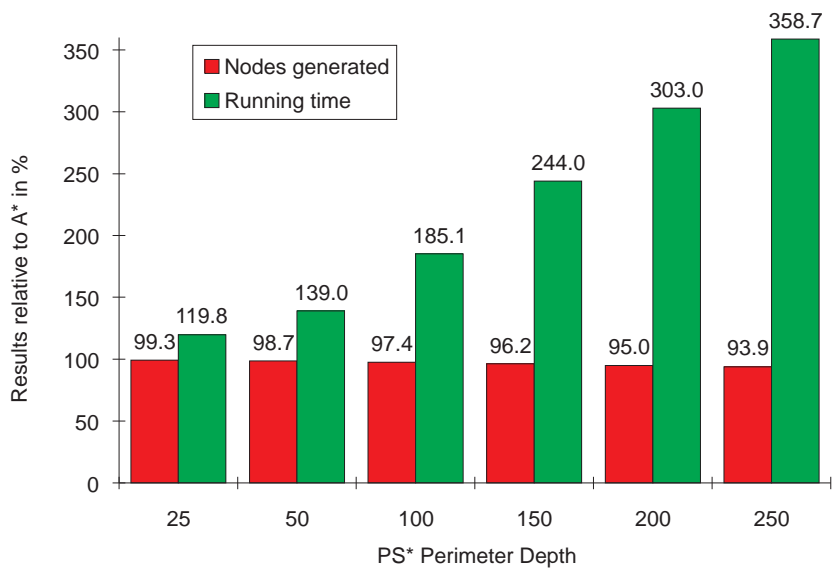

Figure 5: Comparison of PS* for different perimeter depths on the maze problems (100 instances).

Puzzle just contains two nodes. Still, the perimeter approach saves about half of the node generations of IDA*.

This major improvement can be explained quite simply when looking at an approach to improving the heuristic evaluation function. Perimeter search "discovers" during the search an analogous improvement of the Manhattan distance heuristic to that presented by Korf and Taylor (1996, p. 1203) under the name "last moves heuristic" (more precisely the part dealing exactly with the very last move).[8] More precisely, in most cases the dynamic values increase $h_1(n)$ by two units, i.e., twice the (unit) cost of either of the arcs from the two perimeter nodes.[9] Through such improved evaluations, many node generations can be saved even when using just very few perimeter nodes.

Still, the question remains why such improvements are not observed in the maze domain. While in both domains the arcs have unit costs, we found some major differences that help us explain this phenomenon. Fifteen Puzzle problems have relatively short (optimal) solutions, and due to the unit costs of the arcs the overall cost of a solution is also relatively small (53.1 on average). In comparison, maze problems (in mazes of the size we used) have relatively long (optimal) solutions and relatively high cost of such a solution (5262 on average). These differences are also reflected in differences of the heuristic values (although we used in both domains more or less the same heuristic). For the given set of Fifteen Puzzle problem instances, $h_1(s) = 37.1$ is on average much smaller than $h_1(s) = 2361$ for the given set of problem instances in the maze domain. While we do not have data on the heuristic values in the "Think-A-Dot" problems as used by Dillenburg and Nelson (1994), note that the mean path length was given there as 18.4, i.e., even much smaller than for the Fifteen Puzzle.

---

8. This heuristic is based on the last move of a solution, which must return the blank to its goal position. In order to allow the blank to this position, those tiles next to the blank in the goal position must be in certain places. If they are not then the Manhattan distance will not accommodate a corresponding path and can therefore be increased by two units.

9. This also relates to a property of the Manhattan distance heuristic itself. In most cases, the increase of the cost through the known arc (with cost 1) is added to an increase of the heuristic estimate from the evaluated node to the perimeter node (also by 1) compared to the estimate to $t$. In the remaining cases, the heuristic estimate from the evaluated node to the perimeter node reduces (by 1) compared to the estimate to $t$, which cancels out the cost through the known arc.

Let us assume now that in both the Fifteen Puzzle and the maze domain with the same number of perimeter nodes twice the cost of an arc (i.e., two units) can be added. This means that the resulting dynamic evaluation improves on the static evaluation by the same *absolute* amount, but by a quite different *relative* amount: 5.4 percent for the Fifteen Puzzle compared to 0.08 percent in the maze domain. So, the dynamic improvement of the heuristic is in effect much higher for the Fifteen Puzzle, which leads to much larger savings in terms of node generations for the same effort through *front-to-front* evaluations.

In summary, for the Fifteen Puzzle just some few perimeter nodes improve the static evaluation, since twice the (unit) costs of their arcs or even more is in most cases simply added. This has a large effect in this domain where the heuristic values are typically smaller than about 40. In the maze instances of the size we experimented with, the heuristic values are two orders of magnitude larger, and therefore many more perimeter nodes would be required to achieve much effect. These, however, make perimeter search very expensive both in terms of running time and probably also in storage requirement.

From these considerations, it should be clear that the effect of *front-to-front* evaluations is not so much steering the frontiers together, but rather to improve the heuristic evaluations dynamically. In particular, the example of having just two perimeter nodes illustrates both that "wave shaping" is not the real effect, but rather improvement of evaluation accuracy.

## 4. A Generic Approach to Non-Traditional Bidirectional Search

We developed a new generic approach to bidirectional heuristic search that integrates various search algorithms and typically leads to hybrid combinations. Since this approach does not allow for changing the search direction more than once, it can be viewed as a non-traditional form of bidirectional search.

The major steps of our *generic approach* are:[10]

1. Assign a search direction and some or even nearly all of the available memory to the traditional best-first search.

2. Perform traditional best-first search in the assigned direction using the given memory.

3. Unless the best-first search has already found an optimal solution, perform a search in the reverse direction. Use the memory structure built up by the previous best-first search, possibly together with additional memory that is still available, but compute and use *front-to-end* evaluations.

It would not be too difficult to perceive an even more general approach that subsumes perimeter search. Because of the expensive *front-to-front* evaluations, however, we wanted to devise an approach that avoids the need to find a balance between the cost of such evaluations and their beneficial effect.

---

10. This approach is different from the one that we proposed earlier (Kaindl, Kainz, Leeb, & Smetana, 1995).

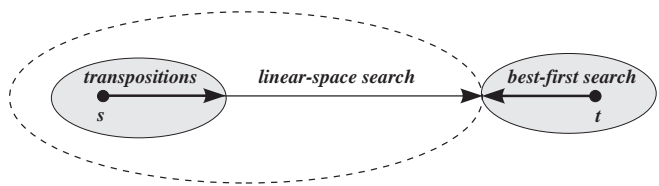

Figure 6: A specialization of our generic approach.

A useful *specialization* of our generic approach that uses memory on both sides of the search space is illustrated in Fig. 6. The traditional best-first search uses its assigned memory as usual, e.g., in A*, and the linear-space search uses as much memory as still available in a transposition table (Reinefeld & Marsland, 1994). The former first of all orders the sequence of node generations and finds transpositions. The latter uses its memory for finding transpositions in another part of the search space, and for caching more accurate heuristic evaluations closer to $t$.

When limited memory is available, this approach is very flexible. For instance, when no memory for a transposition table is assigned, this approach combines linear-space search with conventional best-first search in a bidirectional style. While this may look quite similar to BIDA*, note that our approach in contrast performs *front-to-end* evaluations. The memory of the best-first search is used to find solutions earlier by meeting its frontier (rather than $t$).

When sufficient memory is available even for solving the most difficult problem instances in a domain, also the search in the reverse direction may be performed as a traditional best-first search like A*. After all, A* is under certain conditions and in a certain sense optimal with respect to node expansions (Dechter & Pearl, 1985).

## 4.1 Instantiating for Limited Memory

First we show how our generic approach can be instantiated when only limited memory is available. Of course, any such instantiation should make use of any available domain-specific information. In particular, it should combine those unidirectional search algorithms that best suit the properties of the domain (see, e.g., Rao et al., 1991; Zhang & Korf, 1993). For example, in some domains IDA* is the choice, while in others *depth-first branch-and-bound* (Lawler & Wood, 1966) is much better. In the case of limited memory, either of them is to be preferred over A*.

Below we will present experimental results on the Fifteen Puzzle, a domain that is characterized by having only few distinct cost values. Under this condition, it is reasonable to select IDA* as a linear-space search algorithm, since difficult problem instances of the Fifteen Puzzle require too much memory using A*, when only the Manhattan distance heuristic is used. Since A* makes good use of consistent heuristics like this one (Dechter & Pearl, 1985), we select it for the part of the best-first search.

Based on the key idea of bidirectional search, we let A* and IDA* search in opposite directions in steps 2 and 3 of our generic approach, respectively. This instantiation of our generic approach leads to BAI (Bidirectional A* – IDA*).

Optionally, we may also give the IDA* search some part of the available memory as a transposition table. Fig. 6 illustrates this instantiation. We call this variant of BAI due to the use of this table BAI-Trans.

If A* cannot find a solution using the given memory, then IDA* searches in the reverse direction towards the frontier of the prior search. Since we consider the case of finding *optimal* solutions, this search cannot always terminate immediately after a solution is found. A better solution may exist, and the algorithm must find an optimal one and subsequently prove that it is optimal.

More technically, the IDA* part must be changed slightly. Instead of having to find the goal node, a solution is found whenever the depth-first search meets the frontier of the opposing A* search. If the cost of this solution is smaller than the cost of the best solution found so far (or if it is the first solution found) then its value is stored. Of course, the cost of the best solution found so far may be sub-optimal, or the algorithm does not yet know that it is already optimal. However, if the stored value does not exceed the non-overestimating threshold of the IDA* part, then its depth-first search is exited successfully with an optimal solution.

In addition to these necessary changes, the IDA* part has the advantage to start with an increased initial threshold based on an *admissible* estimate of the optimal solution cost as determined by the A* part. Since we assume a consistent heuristic $h$, the minimum of $f = g + h$ for all nodes in OPEN is always an admissible estimate. Therefore, if this estimate is higher than the usual initial threshold of IDA*, then it can be used here instead.

Moreover, it is not necessary to have the IDA* part search again in the space already explored by A*. More technically, when the depth-first search invoked by IDA* meets a closed node of the opposing A* search frontier, this branch can be cut off (meeting an open node is in general insufficient). We call this *nipping* according to the analogous method described by Kwa (1989).

In an efficient implementation of the Fifteen Puzzle even the effort of hashing at every node causes an overhead that cannot be ignored. Therefore, we implemented BAI in such a way that it avoids hashing at those nodes where — based on the heuristic estimate — it knows that the frontier of the opposing A* search is still out of reach.

According to step 1 of our generic approach, the search directions must be assigned to the A* and the IDA* part, respectively. For traditional bidirectional search, Pohl (1971) proposed and used a *cardinality criterion* for the problem of determining the frontier from which to select a node for expansion: continue searching from the frontier with fewer open nodes. While this is utilized for each node expansion in traditional bidirectional search algorithms, BAI has to decide this issue once at the very beginning of the whole search.

When the search space is sufficiently symmetric, the initial search direction can be determined at random. When the search space is at least slightly asymmetric and no specific knowledge for determining the search direction is available, it seems reasonable to make shallow probes into the search space from both sides and to use the idea behind the cardinality criterion. Since BAI incorporates IDA*, using this algorithm also for *probing* is consistent with the overall approach. For example in the Fifteen Puzzle, the first few iterations of IDA* are searched from both sides, and the direction with fewer generated nodes is assigned to the IDA* part of the overall search, since especially for difficult problems it will have to search much deeper than the A* part.

Let us shortly discuss the behavior of BAI. In the *best case*, it would seem to be the same as A*. In fact, BAI can even be better than pure A*. BAI assigns the search direction dynamically, which can lead to better results than systematically going in one direction. In the *worst case*, BAI has to perform the part of A*, without savings in the IDA* part (except the effect of nipping).

A key question is how BAI saves effort without having enough memory available for completing the A* search. Primarily, it can save one or more of IDA*'s iterations. Due to the better initial threshold, some of the early iterations can be saved. Since the earlier iterations are comparably cheap, this helps much less than saving the last iteration. The search can also be terminated *after a complete* iteration of IDA* if the cost of the best solution already found is not larger than the new increased threshold. Therefore, large savings are possible when BAI terminates earlier than pure IDA*.

## 4.2 Instantiating for Sufficient Memory

Now let us sketch how our generic approach can be instantiated for the case that sufficient memory is available in the sense that even for solving the most difficult problem instances in a domain, traditional best-first search can terminate successfully with the given memory. This case is of interest in order to see whether bidirectional search can be better than A*, which is in some sense optimal over unidirectional algorithms.

When sufficient memory is available, instead of IDA* (or depth-first branch-and-bound) the reverse search can employ A*. In fact, it is easy to construct such an algorithm analogously to BAI as described above, just by using A* instead of IDA*. This instantiation of our generic approach leads to BAA (Bidirectional A* − A*). This algorithm changes the search direction only once in contrast to BS*. For a better utilization of our approach to dynamically improving heuristic values based on differences, we will introduce a slight variation of this algorithm below.

## 5. An Approach to Dynamically Improving Heuristic Values based on Differences

Our new approach to dynamic improvements of heuristic evaluations during search is based on differences between known costs and heuristic estimates. Such differences are utilized by two concrete methods as presented below. The basic idea common to these methods is that for many nodes during the search, the actual cost of a path to or from them is already known. Since the static heuristic values can normally be gained rather cheaply, the *differences* can be computed that signify the error made in their evaluation compared to the cost of the known path. These differences are utilized to improve other heuristic estimates during the same search.

In order to be able to compute such differences, the search must be bidirectional. We focus here on the context of our non-traditional approach to bidirectional heuristic search described above. Actually, application is also possible in the context of traditional bidirec-tional search like BS*. This involves, however, intricacies that are beyond the scope of this paper. So, the interested reader is referred to (Kainz, 1996).

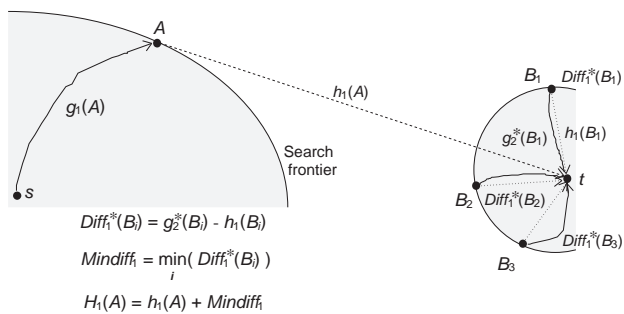

Figure 7: An illustration of the *Add* idea.

## 5.1 The *Add* Method

The first method instantiates this approach by *adding* a constant derived from such differences to the heuristic values of the static evaluation function. Therefore, we call it the *Add* method.

Note, that adding a constant to all evaluations does not change the order of node expansions in a unidirectional search algorithm like A*. So, the benefit from this approach may not be immediately obvious. However, in bidirectional search algorithms using *front-to-end* evaluations, estimates are compared to the cost of the best solution found so far (which is not necessarily already an optimal one), and having better estimates available for such comparisons improves the efficiency due to earlier termination. We explain in more detail below how we apply this approach in the context of our non-traditional approach to bidirectional heuristic search.

See Fig. 7 for the key idea of this method. We assume consistency of the static heuristic evaluator $h_d$. Around the goal node $t$, a search has examined a part of the graph and stored all the optimal paths from nodes $B_i$ on its closed fringe to $t$. For each node $B_i$, its heuristic value $h_1(B_i)$ is computed and subtracted from the optimal path cost $g_2(B_i) = g_2^*(B_i) = h_1^*(B_i)$, resulting in $Diff_1^*(B_i)$. This is actually the error made by the heuristic evaluation of node $B_i$. The minimum of $Diff_1^*(B_i)$ for all nodes $B_i$ on the fringe is computed — we call it $Mindiff_1$.

The point of the *Add* method is that any consistent heuristic value $h_1(A)$ for some node $A$ outside this stored graph underestimates $h_1^*(A)$ by *at least $Mindiff_1$*. We prove this precisely below, but first we need to show a result about $Diff_1^*$.

**Lemma 5.1** *If the heuristic $h_1$ is consistent, then on any optimal path from some node $n$ to $t$ with an intermediary node $m$*

$$Diff_1^*(m) \leq Diff_1^*(n)$$

*holds, i.e., $Diff_1^*$ can only decrease on an optimal path with decreasing distance to the goal node $t$.*

**Proof:** If the heuristic $h_1$ is consistent, then we have

$$h_1(n) \leq h_1(m) + k_1(n,m)$$

From this we simply obtain

$$g_2^*(m) - h_1(m) \leq g_2^*(m) + k_1(n,m) - h_1(n)$$

Since $n$ and $m$ are on one optimal path to $t$, we know that

$$g_2^*(n) = g_2^*(m) + k_1(n,m)$$

After substitutions we obtain

$$g_2^*(m) - h_1(m) \leq g_2^*(n) - h_1(n)$$

and equivalently

$$Diff_1^*(m) \leq Diff_1^*(n)$$

which proves the lemma. □

**Theorem 5.1** *If the heuristic $h_1$ is consistent, then it is possible to compute an admissible heuristic $H_1$ for some node $A$ outside the search frontier around $t$ by*

$$H_1(A) = h_1(A) + Mindiff_1 \leq h_1^*(A)$$

**Proof:** When some path exists from node $A$ to $t$, also an optimal path must exist, and let it go through the frontier node $B_j$. (If no such path exists, $h_1^*(A)$ is infinite and the theorem holds.) From Lemma 1 and the definition of $Mindiff_1$ we know that

$$Mindiff_1 \leq Diff_1^*(B_j) \leq Diff_1^*(A)$$

Since $Diff_1^*$ is the error made by the heuristic $h_1$, we can write

$$h_1(A) + Diff_1^*(A) = h_1^*(A)$$

After substitution we obtain

$$h_1(A) + Mindiff_1 \leq h_1^*(A)$$

which proves the theorem. □

**Corollary 5.1** *$H_1(A)$ is also an admissible estimate if $A$ is a frontier node.*
**Proof:** We can replace $A$ by $B_j$ in the proof of Theorem 3.1 without changing its validity. □

**Theorem 5.2** *If the heuristic $h_1$ is consistent, then $H_1$ is consistent.*
**Proof:** If the heuristic $h_1$ is consistent, then we have

$$h_1(n) \leq h_1(m) + k_1(n, m)$$

Adding the constant $Mindiff_1$ on both sides leads to

$$h_1(n) + Mindiff_1 \leq h_1(m) + Mindiff_1 + k_1(n, m)$$

This means that

$$H_1(n) \leq H_1(m) + k_1(n, m)$$

which proves the theorem.                                        □

Now let us sketch how this *Add* method can be utilized in the context of our non-traditional approach to bidirectional heuristic search. When using it in BAA, for example, the first A\* search must be used to compute some value $Mindiff_1$ (we assume that it starts from node $t$). Optimal paths to all nodes within the search frontier are guaranteed but not to all frontier nodes themselves. If a suboptimal path was found to some frontier node, however, it is known that an optimal path leads through another frontier node with an optimal path to $t$. So, this does not change $fmin$, since the costs of suboptimal paths cannot influence the minimum. $Mindiff_1$ is during the reverse A\* search just a constant to be added to $h_1$. We call the resulting algorithm Add-BAA.

Of course, a larger value of $Mindiff_1$ is to be preferred for a given amount of search. So, the search starting around $t$ should be better guided by expanding always one of those nodes $n$ with minimal $Diff_1^*(n)$. We call this variant here Add-BDA.[11]

It is also necessary to check, whether a node to be evaluated is outside or on the fringe of the graph around $t$. This is simply achieved in Add-BAA and Add-BDA through hashing, which is to be done anyway. When a node on the fringe of the first A\* search is matched, a solution is already found, and when the first node of a path inside the stored graph around $t$ is matched, this path need not be pursued any further, since its optimal continuation is already known. So, only the evaluator $H_1$ is actually used, which is consistent, and therefore A\* does not have to re-open nodes (Pearl, 1984). The search terminates when it selects some node $n$ for expansion with $f_1(n) = g_1(n) + H_1(n)$ not being smaller than the cost of the best solution found so far, which is proven this way to be an optimal one.

For more details on this method and its theoretical properties we refer the interested reader to (Kainz, 1994).

## 5.2 The *Max* Method

The second method computes its own estimate based on such differences and uses the *maximum* of this and the static estimate. Therefore, we call it the *Max* method.

See Fig. 8 for the key idea of this method. We assume consistency of the static heuristic evaluator $h_d$, and that a path from $s$ to $A$ with cost $g_1(A)$ is known. So we know for its evaluation of node $A$: $h_2(A) \leq g_1(A)$. The difference is $Diff_2(A) = g_1(A) - h_2(A)$. We use this difference for the construction of an admissible estimate $F_1(A)$ of the cost of an optimal path from $s$ to $t$ that is constrained to go through $A$. Note, that $g_1(A) = g_1^*(A)$ is *not* necessary, so we call the difference used here $Diff_2(A)$ instead of $Diff_2^*(A)$.

---

11. Earlier we called it Add-A\* (Kainz & Kaindl, 1996).

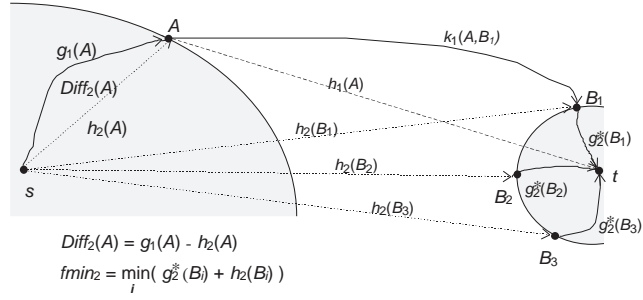

Figure 8: An illustration of the $Max$ idea.

In addition, we assume that a search has been performed from $t$ in the reverse direction. From this search, we assume that from all nodes $B_i$ on its closed fringe optimal paths to $t$ are known, with cost $g_2^*(B_i)$. Therefore, it is possible to compute

$$fmin_2 = \min_i(g_2^*(B_i) + h_2(B_i))$$

Based on these assumptions, we can again construct a dynamic evaluation function as follows.

**Theorem 5.3** *If the heuristic $h_1$ is consistent, then it is possible to compute an admissible heuristic $h_1'$ for some node $A$ outside the search frontier around $t$ by*

$$h_1'(A) = fmin_2 - h_2(A) \leq h_1^*(A)$$

**Proof:** Every path from $A$ to $t$ must go through some frontier node $B_j$. The cost $C_j$ of any such path is bounded from below as follows:

$$C_j \geq k_1(A, B_j) + g_2^*(B_j)$$

If $h_1$ is consistent, it is possible to estimate the optimal cost of a path between two nodes through

$$k_1(A, B_j) \geq h_2(B_j) - h_2(A)$$

Therefore, we can write

$$C_j \geq h_2(B_j) - h_2(A) + g_2^*(B_j)$$

Since $fmin_2 = \min_i(g_2^*(B_i) + h_2(B_i))$, we can also write

$$C_j \geq fmin_2 - h_2(A)$$

This is valid for the cost of any path from $A$ to $t$ including an optimal one, and so we can conclude

$$h_1^*(A) \geq fmin_2 - h_2(A)$$

which proves the theorem. □

**Corollary 5.2** $h'_1(A)$ *is also an admissible estimate if $A$ is a frontier node.*
**Proof:** We can replace $A$ by $B_j$ in the proof of Theorem 3.3 without changing its validity.
□

This dynamic evaluation function is not necessarily better for all nodes than the static function, and so it is useful to combine these functions:

$$H_1(A) = \max(h_1(A), fmin_2 - h_2(A))$$

Since both are admissible the resulting function is also admissible. When the value $fmin_2$ changes during the search, however, $H_1$ is not consistent.

Since in the formula for computing $H_1$ the originally derived difference $Diff_2(A) = g_1(A) - h_2(A)$ is not included, we also derive here the overall evaluation function

$$F_1(A) = \max(f_1(A), fmin_2 + Diff_2(A))$$

Now let us sketch how this *Max* method can be utilized in the context of our non-traditional approach to bidirectional heuristic search. When using it in BAI, for example, the A* search starting first must be used to compute some value $fmin_2$ (we assume that it starts from node $t$). Again, like in the *Add* method, it is not necessary that optimal paths from $t$ to all frontier nodes are known. For getting values $fmin_2$ that are as large as possible for a given amount of search, the usual strategy of selecting a node with minimal $f_2$ is appropriate here.

The subsequent IDA* search within BAI must perform hashing in the graph stored around $t$ in order to check, whether a node to be evaluated is outside or on the fringe of the graph around $t$. In the latter case a new solution is found. We call the resulting algorithm Max-BAI. When a transposition table (Reinefeld & Marsland, 1994) is used in addition as in BAI-Trans, we call it Max-BAI-Trans.

Most interestingly, IDA* can also utilize the *Max* method *without additional storage requirements*. Let us sketch the basic approach for such a *linear-space* application of this method here. While IDA* normally searches in one direction only, we let it alternate the search direction after each iteration until a solution is found. Actually, this procedure is outside our generic approach to bidirectional search as presented above. But we include it here since a linear-space approach is of special interest. $fmin_i$ is computed in one iteration to be used in the subsequent iteration, which must search in the alternate direction so that it can use this value. For example, in an iteration searching from $t$ to $s$, the adapted IDA* computes $hmax_1 = \max(h_1(B_i))$ for all nodes $B_i$. This value is used as an estimate in the subsequent iteration for checking, whether a node $A$ to be evaluated is "outside": if $h_1(A) > hmax_1$ is true, then node $A$ cannot be "inside" and $H_1(A)$ can be safely used. This check substitutes hashing in a stored graph. Since the static heuristic function normally underestimates, however, for some nodes the heuristic $H_1$ is not used although it would theoretically be correct to use it. We call the resulting algorithm that is based on this idea Max-IDA*.

For more details on this method and its theoretical properties we refer the interested reader to (Kainz, 1996).

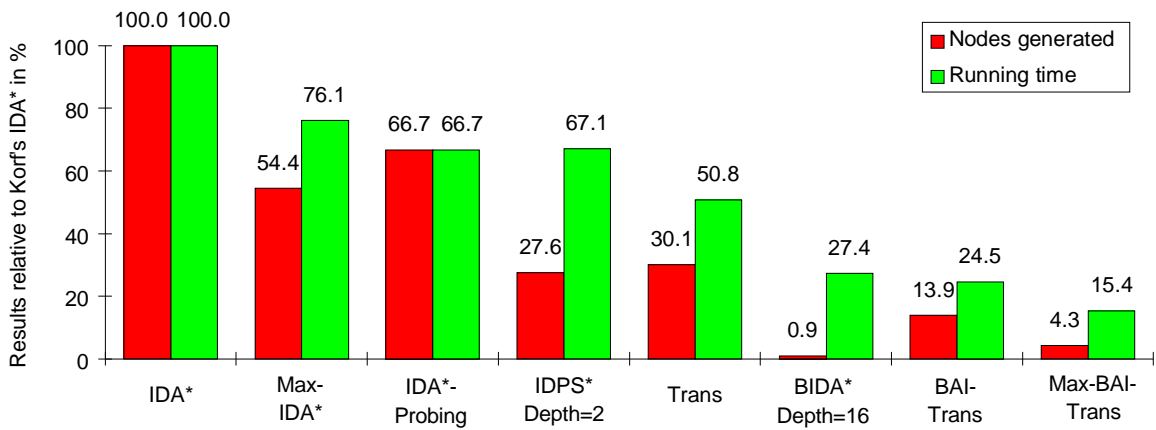

Figure 9: Comparison on the Fifteen Puzzle (100 instances).

## 6. Results of Experiments with these New Approaches

In order to provide some empirical evidence for the effectiveness and the efficiency of our new approaches, we made experiments in two different domains: Fifteen Puzzle and mazes.

### 6.1 Fifteen Puzzle

First let us have a look on specific experimental results for finding optimal solutions to a set of Fifteen Puzzle problems, once again the complete set of 100 instances used by Korf (1985). We compare algorithms that achieve the previously best results in this domain with our new algorithms. All the compared algorithms use no domain-specific knowledge about the puzzle other than the Manhattan distance heuristic.[12] The main storage available on the Convex C3220 used was up to 256 Mbytes.

Fig. 9 shows a comparison of several algorithms in terms of the average number of node generations and their running times. The data are normalized to the respective search effort of IDA* (in Korf's implementation). As already noted above, IDA* needs on average slightly less than half an hour on the machine that we used to find an optimal solution to one problem instance. So, even slight improvements mean notable savings in time.

IDA*, Max-IDA* and IDA*-Probing are linear-space algorithms that use no additional storage, and so their performance cannot compete with the algorithms that use up to 256 Mbytes. Max-IDA* generates only 54.4 percent of the number of nodes generated by IDA* due to its dynamic improvements of the heuristic evaluations according to our *difference* approach. Since these, however, imply some overhead per node searched, it needs 76.1 percent of IDA*'s running time. IDA*-Probing is a variant of IDA* that just uses our *probing* idea for selecting the search direction. Although the search space of the sliding-tile puzzle appears to be quite symmetric, it is interesting to see how much can be gained here just by selecting the search direction dynamically. Since IDA*-Probing has no overhead in running time, it is even faster than Max-IDA*. In order to see how well probing via three iterations already indicates the better search direction, we compared its result with that of a perfect oracle. Using it would still generate 64 percent of IDA*'s nodes, i.e., IDA*-Probing with an overhead in generated nodes for determining the search direction of only less than 0.1 percent is overall just 3 percent worse than this. Systematically searching in the backward direction, however, is not significantly better than systematically searching in the forward direction due to high standard deviations, although it saves 17 percent.

---

12. With much improved heuristic functions, much more efficient searches result (Culberson & Schaeffer, 1996) and even solving Twenty-Four Puzzle instances has become feasible (Korf & Taylor, 1996).

IDPS* uses just some few nodes of additional storage for its perimeter. Due to the related overhead of the *front-to-front* evaluations, it needs about the same running time as IDA*-Probing, although it generates much fewer nodes.[13]

Trans (using 256 Mbytes of memory) achieves savings of about half of the running time compared to IDA*. It saves even much more node generations with this amount of memory, but the effort for hashing slows it down.[14]

Another technique to prune duplicate nodes was proposed by Taylor and Korf (1993), using a finite state machine. Its results are not included in Fig. 9, since we lack data on the running time (no such data are given by Taylor and Korf (1993), and we did not re-implement this technique). IDA* employing this pruning technique generated 100.7 million nodes on the same set of instances as reported by Taylor and Korf (1993), which means 27.7 percent of the number of nodes generated by pure IDA*. The finite state machine that achieved this result contained 55,441 states, requiring only a modest amount of storage. Of course, the finite state machine must be built in a pre-processing stage first. But its use during the search involves only a small and constant overhead in running time. So, for the sliding-tile puzzles, this approach seems to be better than transposition tables for eliminating duplicates. It actually appears to represent the most successful approach yet to solving Fifteen Puzzle problems using unidirectional search.

In principle, we have provided all the available storage to BIDA* (Manzini, 1995), the most efficient algorithm of the perimeter approach. In the given 256 Mbytes of storage, BIDA* can store a maximum of 1 million perimeter nodes. This would correspond to a perimeter depth of 19, where BIDA* generates just 0.4 percent of the number of nodes generated by IDA*, but needs 42 percent of IDA*'s running time. So, as shown in Fig. 4 above it can use more memory for further savings in the number of nodes generated, but it has an optimum in running time for a smaller perimeter size (16), that we show in Fig. 9. Also with the reduced perimeter, BIDA* achieves the best result in terms of nodes

---

13. The results reported by Dillenburg and Nelson (1994) are based on runs using a different sample set of the Fifteen Puzzle, and a different perimeter depth. Using the same perimeter depth (4), the results on Korf's set with our re-implementation are even better in terms of the number of node generations, but very much slower in terms of running time (even slower than IDA*). In personal communication with John Dillenburg it turned out that their implementation of IDA* is slower than Korf's one (which we are using) by a factor of about 60 per generated node. In such an implementation the overhead especially of wave shaping does not show up that clearly as it does in an efficient one. Since smaller perimeter depth means fewer stored nodes and therefore less overhead through wave shaping, the perimeter depth 2 results in better running time, and consequently we show these data in our figure.

14. The data in the figure were gained using a re-implementation of Trans based on efficient code provided by Jonathan Shaeffer. Note the different way of presenting the results: *absolute* data in our figure vs. *relative* to problem difficulty by Reinefeld and Marsland (1994). We had to re-implement Trans, since no data about the performance of Trans with the amount of memory that we used were available, and since we integrate this technique into some of our algorithms. Actually, Trans+Move is the best algorithm described by Reinefeld and Marsland (1994), but its *absolute* results are less than one percent better than those of Trans. Therefore, we did not re-implement Trans+Move and cannot include it into the figure.

generated — just 0.9 percent of the number of nodes generated by IDA*. While BIDA*'s overhead for computing *front-to-front* evaluations is smaller than that of IDPS*, BIDA* needs 27.4 percent of IDA*'s running time.[15]

Our algorithms BAI-Trans and Max-BAI-Trans can store a maximum of 5 million nodes in our implementation of these algorithms in the given 256 Mbytes of storage. BAI-Trans generates clearly more nodes (13.9 percent of IDA*) than BIDA*, but since its overhead per node is much smaller, its running time is even slightly better (24.5 percent). Max-BAI-Trans — additionally utilizing our new *difference* approach — achieves the fastest searches, needing just 15.4 percent of the time needed by IDA*. For achieving this result, it uses 4 million nodes for the *Max* method (and BAI) and 1 million nodes for Trans. In order to see the influence of Trans, we compare this result with that of Max-BAI (not shown in Fig. 9 in order not to clutter it) that uses just the 4 million nodes for the *Max* method (and BAI). Needing 19.2 percent of the time used by IDA*, it is just slightly slower than Max-BAI-Trans, which shows the comparably modest influence of Trans.

In summary, our new approach to bidirectional heuristic search enhanced by our *Max* method achieves the fastest searches for finding optimal solutions on the Fifteen Puzzle of all those using the Manhattan distance heuristic as the only knowledge source. The superiority of Max-BAI-Trans in terms of running time over previous algorithms is statistically significant. For example, the probability that the improvement of the running time over BIDA* is due to chance fluctuation is smaller than 0.15 percent according to a test that compares the means of the paired samples of the absolute running times, and it is even much smaller according to the same test for the data relative to the difficulty of each instance as well as according to the sign test.[16] When using less efficient implementations of IDA* as the basis, the difference would become smaller, since our approach has less overhead per node searched and therefore "gains" less compared to pure IDA*. However, we prefer to compare the algorithms using the most efficient implementation that we have available. For more details on these results see (Kainz, 1996).

## 6.2 Mazes

In order to get a better understanding of the usefulness of our new approach, we made also experiments in a second domain — finding shortest paths in a maze. These are the same maze problems as described above in Subsection 3.2. In addition to these $2000 \times 2000$ mazes, we also made experiments with much smaller $1000 \times 1000$ mazes, in order to see whether the size influences the relative performance of the various algorithms. We compare known algorithms that achieve the best results in this domain (as far as we found) with our algorithms Add-BAA and Add-BDA. The traditional shortest-path algorithm by Dijkstra (1959) corresponds to A* without using heuristic knowledge, so we need not explicitly include it in our experiments. Also in these experiments, all the compared algorithms use no domain-specific knowledge other than the Manhattan distance heuristic, and the main storage available on the Convex C3220 used was up to 256 Mbytes.

---

15. As noted already above in Subsection 3.2, BIDA*'s result here is worse than the data reported by Manzini (1995), which is primarily due to using a different machine and a different implementation that is based on the very efficient code of IDA* for the puzzle provided to us by Korf that we are using.

16. For more details on the statistic tests used we refer the interested reader to (Kaindl, Leeb, & Smetana, 1994; Kaindl & Smetana, 1994).

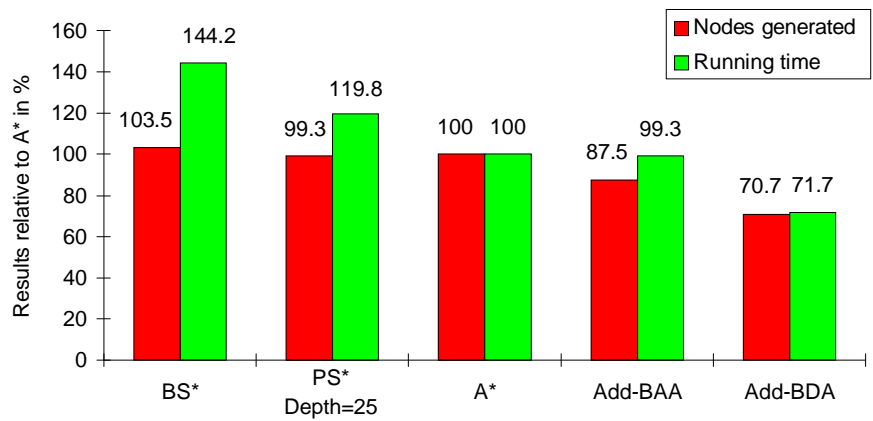

Figure 10: Comparison on the maze problems (100 instances).

Fig. 10 shows a comparison of several algorithms in terms of the average number of node generations and their running times. The data are normalized to the respective search effort of A*. As already noted above, A* needs on average less than two minutes on the machine that we used to find an optimal solution to one problem instance.

BS* generates slightly more nodes for solving these problems than A* (103.5 percent), and its running time is even worse. While it may seem that the implementation of BS* could be further optimized, it is clear that there is some overhead as compared to A*. So, BS* can certainly not improve on A* here.

PS* (Dillenburg & Nelson, 1994) — using perimeter search, i.e., the *front-to-front* method — generates 99.3 percent of the number of nodes of A*, but it needs 119.8 percent of the time used by A*. These data correspond to a perimeter depth of 25, and they are the best results of those shown in Fig. 5 above in terms of running time (see also the discussion in Subsection 3.2). So, also PS* cannot really improve on A* here.

Our algorithms Add-BAA and Add-BDA generate clearly fewer nodes than A* (87.5 and 70.7 percent, respectively). The better performance of Add-BDA reflects the higher $Mindiff_1$ value that is achieved through guiding the first of the two best-first searches by expanding always one of those nodes $n$ with minimal $Diff_1^*(n)$. More precisely, Add-BDA achieved $Mindiff_1 = 1174$ (from a reverse search of 750k nodes), while Add-BAA achieved only $Mindiff_1 = 811$ (from a reverse search of even 1000k nodes). The performance of Add-BAA in terms of running time is, however, still much the same as that of A* (at least in this implementation as derived from BS*). Add-BDA achieves the fastest searches, needing just 71.7 percent of the time needed by A*. So, the application of our approach to dynamically improving heuristic values is feasible here with very little overhead.

The superiority of Add-BDA over previous algorithms is statistically significant. For example, the probability that the improvement in terms of running time over A* is due to chance fluctuation is smaller than 0.005 percent according to all the three statistic tests that we made analogously to those for the Fifteen Puzzle data. The same significance result holds for the improvement with respect to the number of node generations. Both Add-BDA and A* as well as the other algorithms compared here generate all child nodes at once in node *expansions*, and the superiority of Add-BDA over these algorithms is statistically significant also in this respect. This is particularly interesting, since the optimality result of A* over unidirectional algorithms is stated in the sense that A* never expands a node that could be skipped by some other (unidirectional) algorithm (Dechter & Pearl, 1985).

Table 1: Overview of approaches to bidirectional heuristic search.

|  | *front-to-front* | *front-to-end* |
|---|---|---|
| traditional | BHFFA, BHFFA2 | BHPA, BS* |
| non-traditional | PS*, IDPS*, BIDA* | Max-BAI-Trans, Add-BDA |

Since the relative results on the $1000 \times 1000$ mazes are very similar, we do not show them explicitly here (see, however, Kainz, 1996). They provide some empirical evidence that the performance of these algorithms is not just peculiar for a certain size of mazes.

## 7. Discussion

After this presentation of our new approach to bidirectional heuristic search and its experimental results, let us put it into perspective. Table 1 provides an overview of the existing approaches according to the way of evaluating and the way of organizing the change(s) of search direction. The algorithms that instantiate our new generic approach fall into the category of non-traditional bidirectional heuristic search algorithms (that change the search direction only once) and that perform *front-to-end* evaluations. While this approach allows coping with limited memory (e.g., in Max-BAI and Max-BAI-Trans), it is also useful in the case of sufficient memory (e.g., Add-BDA).

Due to avoiding expensive *front-to-front* evaluations, our approach to dynamically improving heuristic evaluations is less effective than perimeter search in saving node generations (at least in the Fifteen Puzzle domain). However, it has less overhead and is therefore more efficient per node searched in terms of running time.

From the viewpoint of Table 1, our approach somehow "completes" the picture of bidirectional heuristic search. (Note, however, that the non-traditional approach was found independently of the work on perimeter search.) Still, there should be ample opportunity for further research on bidirectional search, especially when looking at it from other perspectives. Another issue is, e.g., whether linear-space search is involved or not. We propose in this paper Max-IDA*, an algorithm that alternates the search direction before every iteration in order to be able to use information from the previous iteration for improving the heuristic evaluations dynamically. Yet another perspective is whether an algorithm is designed to find optimal solutions or not. In this paper, we only focused on admissible search algorithms. As discussed above, however, there also exist $\varepsilon$-admissible bidirectional search algorithms that guarantee solutions with a known error bound, as well as others that find solutions without any guarantee about their quality (e.g., d-node retargeting).

When contrasting the traditional and the non-traditional approaches to bidirectional heuristic search, it may appear to be strange that the less flexible approach delivers the better results. Why should it be "better" to change the search direction just once? While it is difficult to provide a generally convincing answer to this question, let us summarize some observations:

- Traditional bidirectional search typically requires exponential space. Kaindl and Khorsand (1994) showed that such a search is possible using limited memory, but because of the complexity of such algorithms the runtime efficiency was insufficient.

- For the perimeter depths where perimeter search is successful, its perimeters are much smaller than the frontiers of traditional *front-to-front* algorithms. Through parameterizing the perimeter depth it is possible to balance the effort for *front-to-front* evaluations with their effect of improving heuristic evaluations dynamically.

- The runtime optimizations of BIDA* over IDPS* are only feasible when the perimeter stays constant (at least for each iteration).

- The *Mindiff* value of the *Add* method becomes higher when the search for computing it generates more nodes. So, in the context of a traditional bidirectional search it is initially small.

- Applying the *Max* idea becomes much more complex, e.g., in BS* where both search frontiers change (Kainz, 1996).

In general, one of the major problems of heuristic search is how to use available but limited memory effectively. Pure *unidirectional* approaches to utilizing limited memory led to less convincing results (Chakrabarti et al., 1989; Sen & Bagchi, 1989; Russell, 1992; Ghosh et al., 1994; Reinefeld & Marsland, 1994) than the non-traditional approaches to bidirectional search as shown in Table 1. In particular, our generic approach allows very flexible and effective use of available memory. This is, however, partly due to its integration of various unidirectional strategies. Future work may investigate the direct use of such unidirectional approaches to utilizing limited memory in instantiations of our generic approach to bidirectional search.

In addition, bidirectional search allows the use of memory for dynamically improving heuristic evaluations in ways that are infeasible for strictly unidirectional search. This is demonstrated by the *front-to-front* approach as well as by our difference method. The following simple idea implicitly behind these approaches may further illustrate this. Given a breadth-first (uniform-cost) search to some depth $d$, any node outside its frontier must be at least $d + 1$ steps away from its start $s$. A reverse search towards $s$ may use this fact to compute an estimate for any node outside this frontier that is at least $d + 1$. This idea cannot be used in a strictly unidirectional search. Note, however, that the approaches discussed here are much more complex and useful than this simple idea. Since they take known costs and heuristic estimates as well as differences of these into account, they can provide much better estimates especially for nodes that are far outside the already given opposite search frontier.

In some sense, it is also possible to view our difference approach as *learning*, since also there differences between predicted and actual outcomes are important. Usual machine learning research, however, strives for using the results from one problem instance for solving subsequent instances, which we did not attempt. An in-depth discussion of this relationship is outside the scope of this paper. Note, however, that also the approaches using *front-to-front* evaluations could be considered from this viewpoint.

## 8. Conclusion

Based on new insights about previous approaches to bidirectional heuristic search, we propose in this paper

- a new generic approach to non-traditional bidirectional search with *front-to-end evaluations*, and

- a new approach to dynamically improving heuristic values in this context.

We showed how to successfully instantiate this generic approach for the very important case when available memory is limited. This memory can also be utilized for efficiently improving heuristic values. For certain problems where sufficient memory is available, we proposed an instantiation in the form of an algorithm that challenges A*, which is in a certain sense optimal over unidirectional search algorithms. The optimality result of A* over unidirectional competitors by Dechter and Pearl (1985) does *not* imply that *bidirectional* search cannot be more efficient, and in our experiments we found some empirical evidence that our new algorithm can be more efficient than A* both in terms of node expansions and running time. We also showed that our approach was more efficient in terms of running time than any other bidirectional or unidirectional search approach using the same information in two different domains. These results are statistically significant.

While traditional bidirectional search did not yet achieve improvements over admissible unidirectional search, the non-traditional way of performing the opposing searches in sequence — as exemplified by perimeter search and by our approach — seems to have great potential. In this sense, we show that bidirectional heuristic search is viable and consequently propose that this search strategy be reconsidered.

## Acknowledgements

Over the years, several people cooperated with the first author on research in heuristic search and in particular bidirectional search: Aliasghar Khorsand, Andreas Köll, Angelika Leeb, Harald Smetana and Roland Steiner. Some of their work served as a basis for the work presented in this paper. For our experiments we had a Convex C3220 at the computing center of the TU Vienna available. Our implementations are based on the very efficient code of IDA* and A* for the puzzle made available by Richard Korf and an efficient hashing schema by Jonathan Schaeffer. Finally, we acknowledge the useful comments on earlier drafts by Andreas Auer, Dennis de Champeaux, Stefan Kramer, Giovanni Manzini, Ira Pohl and Roland Steiner. Parts of this paper already appeared in *Proc. Fourteenth International Joint Conference on Artificial Intelligence (IJCAI-95)* and in *Proc. Thirteenth National Conference on Artificial Intelligence (AAAI-96)*.

## Appendix. Glossary of Notation

| | |
|---|---|
| $s, t$ | Start node and goal/target node, respectively. |
| $d$ | Current search direction index; when search is in the forward direction $d = 1$, and when in the backward direction $d = 2$. |
| $C^*$ | Cost of an optimal path from $s$ to $t$. |
| $k_d(m, n)$ | Cost of an optimal path from $m$ to $n$ if $d = 1$, or from $n$ to $m$ if $d = 2$. |
| $g_d^*(n)$ | Cost of an optimal path from $s$ to $n$ if $d = 1$, or from $n$ to $t$ if $d = 2$. |
| $h_d^*(n)$ | Cost of an optimal path from $n$ to $t$ if $d = 1$, or from $s$ to $n$ if $d = 2$. |
| $g_d(n), h_d(n)$ | Estimates of $g_d^*(n)$ and $h_d^*(n)$, respectively. |
| $f_d(n)$ | Static evaluation function: $g_d(n) + h_d(n)$. |
| $f_d^j$ | One of the $f$-values of expanded nodes in search direction $d$. |
| $H_d(n)$ | Dynamic estimate of $h_d^*(n)$. |
| $F_d(n)$ | Dynamic evaluation function: $g_d(n) + H_d(n)$. |
| $L_{\min}$ | Cost of the best (least costly) complete path found so far from $s$ to $t$. |
| $\text{OPEN}_d$ | The set of open nodes in search direction $d$. |
| $\text{CLOSED}_d$ | The set of closed nodes in search direction $d$. |
| $|\text{OPEN}_d|$ | Number of nodes in $\text{OPEN}_d$. |
| $\#(a)$ | Number of nodes expanded by algorithm $a$. |
| $\#_d(a)$ | Number of nodes expanded by algorithm $a$ in search direction $d$. |
| $\#_d^j(a)$ | Number of those nodes with value $f_d^j$ expanded by algorithm $a$ in search direction $d$. |